\documentclass[10pt, conference, compsocconf]{IEEEtran}
\usepackage{graphicx}
\usepackage[utf8]{inputenc}
\usepackage{tabu}
\usepackage{booktabs} 
\usepackage{tabularx} 
\usepackage{adjustbox}

\begin{document}

\title{MLOps - Definitions, Tools and Challenges\\}

\makeatletter

\newcommand{\linebreakand}{%
  \end{@IEEEauthorhalign}
  \hfill\mbox{}\par
  \mbox{}\hfill\begin{@IEEEauthorhalign}}
\makeatother

\author{
    \IEEEauthorblockN{Georgios Symeonidis}
    \IEEEauthorblockA{\textit{ILSP Xanthi’s Division} \\
    \textit{Athena Research and Innovation Centre}\\
    Xanthi, Greece \\
    giorgos.simeonidis@athenarc.gr}
  \and
  \IEEEauthorblockN{Evangelos Nerantzis}
  \IEEEauthorblockA{\textit{MLV Research Group} \\\textit{Dept. of Computer Science} \\
    \textit{International Hellenic University}\\
    Kavala, Greece \\
    e.nerantzis@athenarc.gr}
  \linebreakand 
  \linebreakand 
  \linebreakand 
  \IEEEauthorblockN{Apostolos Kazakis}
  \IEEEauthorblockA{\textit{MLV Research Group} \\\textit{Dept. of Computer Science} \\
    \textit{International Hellenic University}\\
    Kavala, Greece \\
    akazak@gmail.com}
  \and
  \IEEEauthorblockN{George A. Papakostas*}
  \IEEEauthorblockA{\textit{MLV Research Group} \\\textit{Dept. of Computer Science} \\
    \textit{International Hellenic University}\\
    Kavala, Greece \\
    gpapak@cs.ihu.gr}
}

\maketitle
\begin{abstract}
This paper is an overview  of the Machine Learning Operations (MLOps) area. Our aim is to define the operation and the components of such systems by highlighting the current problems and trends. In this context we present the different tools and their usefulness in order to provide the corresponding guidelines. Moreover, the connection between MLOps and AutoML (Automated Machine Learning) is identified and how this combination could work is proposed.
\end{abstract}

\IEEEpeerreviewmaketitle

\section{Introduction}
Incorporating machine learning models in production is a challenge that remains from the creation of the first models until now. For years data scientists, machine learning engineers, front end engineers, production engineers tried to find a way to work together and combine their knowledge in order to deploy ready for production models. This task has many difficulties and it is not easy to overcome them. This is why only a small percentage of the ML projects manage to reach production. In the previous years a set of techniques and tools have been proposed and used in order to minimize as much as possible this kind of problems. The development of these tools had multiple targets. Data preprocessing, models' creation, training, evaluation, deployment, and monitoring are some of them. As the field of AI progresses such kind of tools are constantly emerging.  
\section{Related Work}
MLOps is a relatively new field and as expected there is not much relevant work and papers. In this section we will mention some of the most important and influential work in every task of the MLOps cycle (Figure~\ref{fig:mlops}). At first, Sasu Makineth et al. \cite{sasu} describe the importance of MLOps in the field of data science, based  on  a  survey  where  they  collected  responses from 331 professionals from 63 different countries. As for the data manipulation task, Cedric Renggli et al. \cite{Renggli2021} describe the significance of data quality for an MLOps system while demonstrates how different aspects of data quality propagate through various stages of machine learning development.  Philipp Ruf et al. \cite{Ruf2021} examine the role and the connectivity of the MLOps tools for every task in the MLOps cycle. Also, they present a recipe for the selection of the better Open-Source tools. Monitoring and the corresponding challenges were discussed by Janis Klaise et al. \cite{Klaise2020} using recent examples of production ready solutions using open source tools. Finally Damnian A. Tamburri \cite{Tamburri2020} presents the current trends and challenges, focusing on sustainability and explainability.

\section{MLOps}
MLOps(machine learning operations) stands for the collection of techniques and tools for the deployment of ML models in production \cite{alla2021mlops}. Contains the combination of DevOps and Machine Learning. DevOps \cite{sanjeev} stands for a set of practices with the main purpose to minimize the needed time for a software release, reducing the gap between software development and operations \cite{Fitzgerald2017}\cite{Noah}. The two main principles of DevOps are Continuous Integration (CI) and Continuous Delivery (CD). Continuous integration is the practice by which software development organizations try to integrate code written by developer teams at frequent intervals. So they constantly test their code and make small improvements each time based on the errors and weaknesses that results from the tests. This results in a reduction in the software development process cycle \cite{RAJ2021}. Continuous delivery is the practice according to which, there is constantly a new version of the software under development to be installed for testing, evaluation and then production. With this practice, the software releases resulting from the continuous integration with the improvements and the new features reach the end users much faster \cite{Karamitsos2020}. After the great acceptance of DevOps and the practices of "continuous software development" in general \cite{Fitzgerald2014}\cite{Fitzgerald2017}, the need to apply the same principles that govern DevOps in machine learning models became imperative \cite{alla2021mlops}. This is how these practices, called MLOps (Machine Learning Operations), came about. MLOps attempts to automate Machine Learning processes using DevOps practices and approaches. The two main DevOps principles they seek to serve are: Continuous Integration (CI) and Continuous Delivery (DC) \cite{Noah}. Although it seems simple in reality it is not. This is due to the fact that a Machine Learning model is not independent but is part of a wider software system and consists not only of code but also of data. As the data is constantly changing, the model is constantly called upon to retrain from the new data that emerges. For this reason, MLOps introduce a new practice, in addition to CI and CD, that of Continuous Training (CT), which aims to automatically retrain the model where needed. From the above, it becomes clear that compared to DevOps, MLOps are much more complex and incorporate additional procedures involving data and models \cite{John2021}\cite{Ruf2021}\cite{Treveil2020}{}.    
\begin{figure}
    \centerline{
    \includegraphics[width=\linewidth]{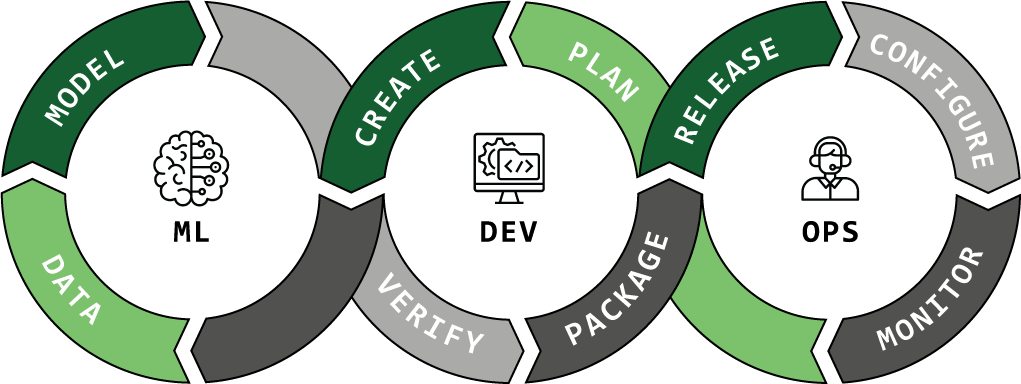}}
    \caption{MLOps Life-cycle.}
    \label{fig:mlops}
\end{figure}
\subsection{MLOps pipeline}
While there are several attempts to capture and describe MLOps, the one that is best known  is the proposal of ToughWorks \cite{git}\cite{tuomas}, which automates the life cycle of end-to-end Machine Learning applications (Figure~\ref{fig:mlops pipeline}). It is "a software engineering approach in which an interoperable team produces machine learning applications based on code, data and models in small, secure new versions that can be replicated and delivered reliably at any time, in short custom cycles". This approach includes three basic procedures involving: collection, selection and preparation of data to be used in model training, in finding and selecting the most efficient model after testing and experimenting with different models, in developing and sending the selected model in production. A simplified form of such a pipeline is shown in Figure \ref{fig:mlops pipeline}.
\begin{figure}
    \centerline{
    \includegraphics[width=\linewidth]{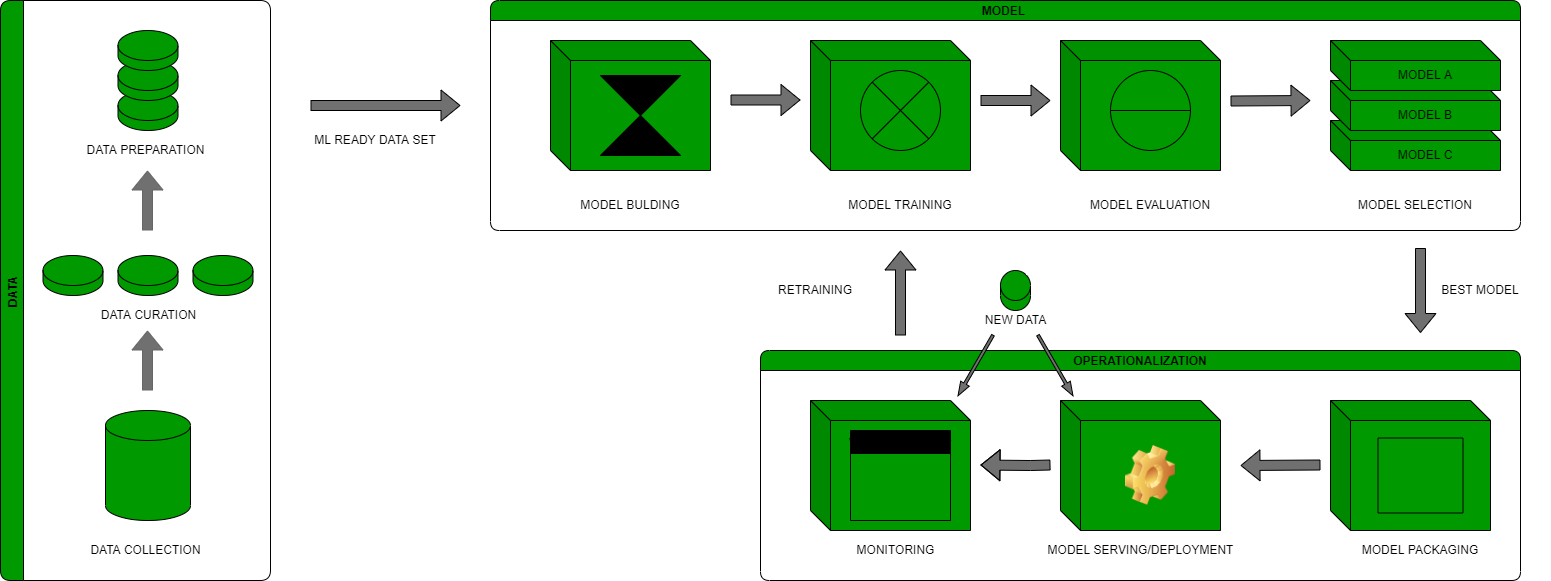}}
    \caption{MLOps Pipeline}
    \label{fig:mlops pipeline}
\end{figure}

After collecting, evaluating and selecting the data that will be used for training, we automate the process of creating models and training them. This allows us to produce more than one model which we can test and experiment in order to produce a more efficient and effective model while recording the results of our tests. Then we have to resolve various issues related to the production of the model, as well as submit it to various tests in order to confirm its reliability before developing it for production. Finally, we can monitor the model and collect the resulting new data, which will be used to retrain the model, thus ensuring its continuous improvement \cite{martin}.

\subsection{Maturity Levels}
Depending on the level of automation of a MLOps system, it can be classified at a corresponding level \cite{John2021}. These levels were named by the community maturity levels. Although there is no universal maturity model, the two main ones were created by Google and Microsoft. Google model consists of three levels and its structure is presented in Figure~\ref{fig:maturitygoogle} \cite{mlop}.
\begin{figure}
    \centerline{
    \includegraphics[width=\linewidth]{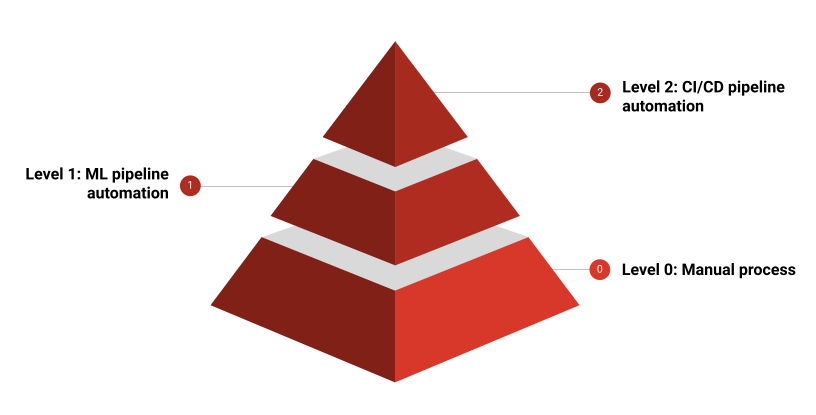}}
    \caption{Googles Maturity Levels.}
    \label{fig:maturitygoogle}
\end{figure}
MLOps level 0: Manual process,
MLOps level 1: ML pipeline automation,
MLOps level 2: CI/CD pipeline automation.
Microsoft model consists of five levels and its structure is presented in Figure~\ref{fig:maturitymicrosoft} \cite{mic}.
Level 1: No MLOps,
Level 2: DevOps but no MLOps,
Level 3: Automated Training,
Level 4: Automated Model Deployment,
Level 5: Full MLOps Automated Operations.
\begin{figure}
    \centerline{
    \includegraphics[width=\linewidth]{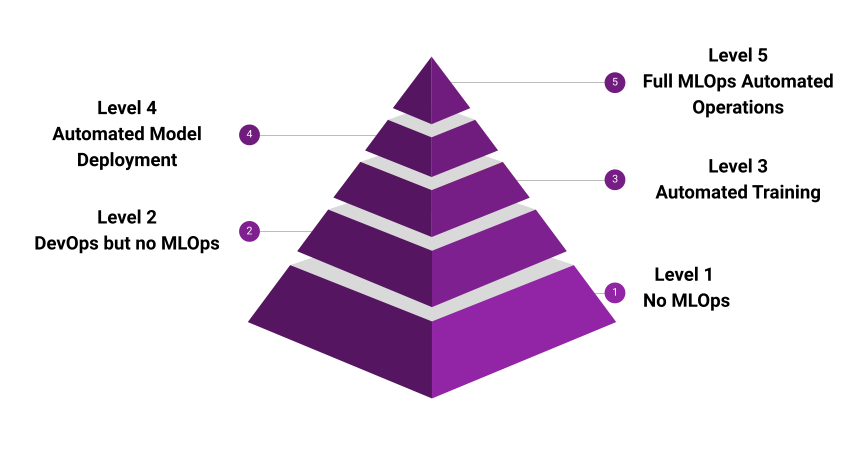}}
    \caption{Microsoft Maturity Levels.}
    \label{fig:maturitymicrosoft}
\end{figure}

\section{Tools and Platforms}
In recent years many different tools have emerged in order to help automate the sequence of artificial learning processes \cite{Felipe2016}. This section provides an overview of the different tools and requirements that these tools meet. Note that different tools automate different phases in the machine learning workflow.
The majority of tools come from the open source community because half of all IT organizations use open source tools for AI and ML and the percentage is expected to be around two-thirds by 2023. At GitHub alone, there are 65 million developers and 3 million organizations contributing to 200 million projects. Therefore, it is not surprising that there are advanced sets of open source tools in the landscape of machine learning and artificial intelligence. Open source tools focus on specific tasks within MLOps instead of providing end-to-end machine learning life-cycle management. These tools and platforms typically require a development environment in Python and R.
In recent years many different tools have emerged which help in automating the ML pipeline. The choice of tools for MLOps is based on the context of the respective ML solution and the operations setup.
\subsection{Data Preprocessing Tools}
Data processing tools are divided into two main categories: \textit{data labeling} tools and \textit{data versioning} tools. Data labeling tools (also called annotation tools, tagging or sorting data), big data labeling plans such as text, images or sound. Data labeling tools can in turn be divided into different categories depending on the task they perform. Some are designed to highlight specific file types such as videos or images \cite{Zhou2017}. Few of these tools can edit all file types. There are also different types of tags that differ in each tool. Boundary frames, polygonal annotations, and semantic segmentation are the most common features in the label market. Your choices about data labeling tools will be an essential factor in the success of the machine learning model. You need to specify the type of data labeling your organization needs \cite{Dietterich2002}. Labeling accuracy is an important aspect of data labeling \cite{Fredriksson2020}. High quality data creates better model performance. Data extraction tools (also called data version controls) by managing different versions of data sets and storing them in an accessible and well-organized way \cite{Armbrust2020}. This allows data science teams to gain knowledge, such as identifying how changes affect model performance and understanding how data sets evolve. The most important  data preprocessing tools are listed in table I.

\begin{table}[h!]
  \begin{center}
    \label{tab:table1}
    \begin{adjustbox}{width=0.5\textwidth}
    \begin{tabular}{c|c|c|c}
      \toprule 
      \textbf{Name} & \textbf{Status} & \textbf{Launched in} & \textbf{Use}\\
      \midrule 
      iMerit&Private&2012&Data Preprocessing\\
      Pachyderm&Private&2014&Data Versioning\\
      Labelbox&Private&2017&Data Preprocessing\\
      Prodigy&Private&2017&Data Preprocessing\\
      Comet&Private&2017&Data Versioning\\
      Data Version Control&Open Source&2017&Data Versioning\\
      Qri&Open Source&2018&Data Versioning\\
      Weights and Biases&Private&2018&Data Versioning\\
      Delta Lake&Open Source&2019&Data Versioning\\
      Doccano&Open Source&2019&Data Preprocessing\\
      Snorkel&Private&2019&Data Preprocessing\\
      Supervisely&Private&2019&Data Preprocessing\\
      Segments.ai&Private&2020&Data Preprocessing\\
      Dolt&Open Source&2020&Data Versioning\\
      LakeFs&Open Source&2020&Data Versioning\\
      \bottomrule 
    \end{tabular}
    \end{adjustbox}
    \caption{Data Preprocessing Tools.}
  \end{center}
\end{table}

\subsection{Modeling Tools}
The tools with which we extract features from a raw data set in order to create optimal training data sets are called \textit{feature engineering} tools. Tools like these have the ability to speed up the feature extraction process \cite{Khalid2014} when applied for common applications and generic problems. To monitor the versions of the data of each experiment and its results as well as to compare between different experiments, we use \textit{experiment tracking} tools, which store all the necessary information about the different experiments because developing machine learning projects involve running multiple experiments with different models, model parameters, or training data.
Hyperparameter tuning or optimization tools automate the process of searching and selecting hyperparameters that give optimal performance for machine learning models. Hyperparameters are the parameters of the machine learning models such as the size of a neural network or types of regularization that model developers can adjust to achieve different results \cite{Bardenet2013}. The most important modeling tools are listed in table II.
\begin{table}[h!]
  \begin{center}
    \label{tab:table2}
    \begin{adjustbox}{width=0.5\textwidth}
    \begin{tabular}{c|c|c|c}
      \toprule 
      \textbf{Name} & \textbf{Status} & \textbf{Launched in} & \textbf{Use}\\
      \midrule 
      Hyperopt&Open Source&2013&Hyperparameter Optimization\\
      SigOpt&Public&2014&Hyperparameter Optimization\\
      Iguazio Data Science Platform&Private&2014&Feature Engineering\\
      TsFresh&Private&2016&Feature Engineering\\
      Featuretools&Private&2017&Feature Engineering\\
      Comet&Private&2017&Experiment Tracking\\
      Neptune.ai&Private&2017&Experiment Tracking\\
      TensorBoard&Open source&2017&Experiment Tracking\\
      Google Vizier&Public&2017&Hyperparameter Optimization\\
      Scikti-Optimize&Open source&2017&Hyperparameter Optimization\\
      dotData&Private&2018&Feature Engineering\\
      Weight and Biases&Private&2018&Experiment Tracking\\
      CML&Open source&2018&Experiment Tracking\\
      MLFlow&Open source&2018&Experiment Tracking\\
      Optuna&Open source&2018&Hyperparameter Optimization\\
      Talos&Open Source&2018&Hyperparameter Optimization\\
      AutoFet&Open Source&2019&Feature Engineering\\
      Feast&Private&2019&Feature Engineering\\
      GuildAi&Open Source&2019&Experiment Tracking\\
      Rasgo&Private&2020&Feature Engineering\\
      ModelDB&Open source&2020&Experiment Tracking\\
      HopsWork&Private&2021&Feature Engineering\\
      Aim&Open source&2021&Experiment Tracking\\
      \bottomrule 
    \end{tabular}
    \end{adjustbox}
    \caption{Modeling Tools.}
  \end{center}
\end{table}

\subsection{Operationalization Tools}
Then to facilitate the integration of ML models in a production environment, we use machine learning \textit{model deployment} \cite{Savu2011} tools. Machine learning \textit{model monitoring} is a key aspect of every successful ML project because ML model performance tends to decay after model deployment due to changes in the input data flow over time \cite{javier}. Model monitoring tools detect data drifts and anomalies over time and allow setting up alerts in case of performance issues. Finally, we should not forget to mention that at this time there are tools available that cover the life cycle of an end-to-end machine learning application. The most important operationalization tools are listed in table III.

\begin{table}[h!]
  \begin{center}
    
    \label{tab:table3}
    \begin{adjustbox}{width=0.5\textwidth}
    \begin{tabular}{c|c|c|c}
      \toprule 
      \textbf{Name} & \textbf{Status} & \textbf{Launched in} & \textbf{Use}\\
      \midrule 
      Google Cloud Platform&Public&2008&end-to-end\\
      Microsoft Azure&Public&2010&end-to-end\\
      H2O.ai&Open source&2012&end-to-end\\
      Unravel Data&Private&2013&Model Monitoring\\
      Algorithmia&Private&2014&Model Deployment / Serving\\
      Iguazio&Private&2014&end-to-end\\
      Databricks&Private&2015&end-to-end\\
      TensorFlow Serving&Open source&2016&Model Deployment / Serving\\
      Featuretools&Private&2017&Feature Engineering\\
      Amazon SageMaker&Public&2017&end-to-end\\
      Kubeflow&Open Source&2018&Model Deployment / Serving\\
      OpenVino &Open source&2018&Model Deployment / Serving\\
      Triton Inference Server&Open source&2018&Model Deployment / Serving\\
      Fiddler&Private&2018&Model Monitoring\\
      Losswise&Private&2018&Model Monitoring\\
      Alibaba Cloud ML Platform for AI&Public&2018&end-to-end\\
      Mlflow&Open source&2018&end-to-end\\
      BentoMl&Open Source&2019&Model Deployment / Serving\\
      Superwise.ai&Private&2019&Model Monitoring\\
      MLrun&Open source&2019&Model Monitoring\\
      DataRobot&Private&2019&end-to-end\\
      Seldon&Private&2020&Model Deployment / Serving\\
      Torch Serve&Open source&2020&Model Deployment / Serving\\
      KFServing&Open source&2020&Model Deployment / Serving\\
      Syndicai&Private&2020&Model Deployment / Serving\\
      Arize&Private&2020&Model Monitoring\\
      Evidently AI&Open Source&2020&Model Monitoring\\
      WhyLabs&Open source&2020&Model Monitoring\\
      Cloudera&Public&2020&end-to-end\\
      BodyWork&Open source&2021&Model Deployment / Serving\\
      Cortex&private&2021&Model Deployment / Serving\\
      Sagify&Open source&2021&Model Deployment / Serving\\
      Aporia&Open source&2021&Model Monitoring\\
      Deep checks&Private&2021&Model Monitoring\\
      \bottomrule 
    \end{tabular}
    \end{adjustbox}
    \caption{Operationalization Tools.}
  \end{center}
\end{table}
\subsection{The example of colossal companies}
It's common for big companies to develop their own MLOps platforms in order to deploy fast and successful, reliable and reproducible pipelines. The main problems that led these companies to create their own platforms are mainly two. Initially, the time needed to build and deliver a model in production \cite{Bosch2021}. The main goal is to reduce the time required, from a few months to a few weeks. Also, the stability of ML models in their predictions and the reproduction of these models in different conditions are always two of the most important goals. Some illustrative examples of such companies are : Google with TFX(2019) \cite{tfx}, Uber with Michelangelo(2015) \cite{michelangelo}, Airbnb with Bighead(2017) \cite{bighead} and Netflix with Metaflow(2020) \cite{metaflow}.

\subsection{How to choose the right tools}
The MLOps life-cycle consists of different tasks. Every task has unique characteristics and the corresponding tools are developing matching with them. Whereat, an efficient MLOps system depends on the choice of the right tools, both for each task and for the connectivity between them. Every challenge also has its own characteristics and the right way to go depends on them \cite{sridhar}. There is not a general recipe one choosing some specific tools \cite{Ruf2021}, but we can provide some general guidelines, that can be helpful at eliminating some tools simplifying this problem. There are tools that offer a variety of functionalities and there are tools that are more specialized. Generally, the fewer tools we use the better because it is easier, for example, to archive compatibility between 3 tools than between 5. But there are some tasks that require better flexibility. So the biggest challenge is to find the balance between flexibility and compatibility. For this reason it is important to make a list of the available tools that are capable of solving the individual problem in every task. Then, we can check the compatibility between them in order to find the best way to go. This requires excellent knowledge of as many tools as possible from every team working on a MLOps system. So the list gets smaller when we add as a precondition the pre-existing knowledge of these tools. This is not always a solution, so we can add tools that are easy to understand and use.

\section{AutoML}
In the last years more and more companies try to integrate machine learning models into the production process. For this reason another software solution was created. AutoML is the process of automating the different tasks that an ML model creation requires \cite{Shubhra}. Specifically, AutoML pipeline contains data preparation, models creation, hyper parameter tuning, evaluation and validation. With these techniques a bunch of models is trained in the same data set,  then a hyper parameter fine tuning is applied, finally the models are evaluating and the best model is exported. Therefore the process of creating and selecting the appropriate model, as well as the preparation of the data, turns into a much simpler and more accessible process \cite{Gijsbers2019}. This is the reason why every year more and more companies turn their attention to AutoML. The combination of AutoML and MLOps simplifies and makes much more feasible the deployment of the ML models in production. In these section we will make a brief introduction into the most modern AutoML tools and platforms aiming at the combination of AutoML and MLOps.
\begin{figure}
    \centerline{
    \includegraphics[width=\linewidth]{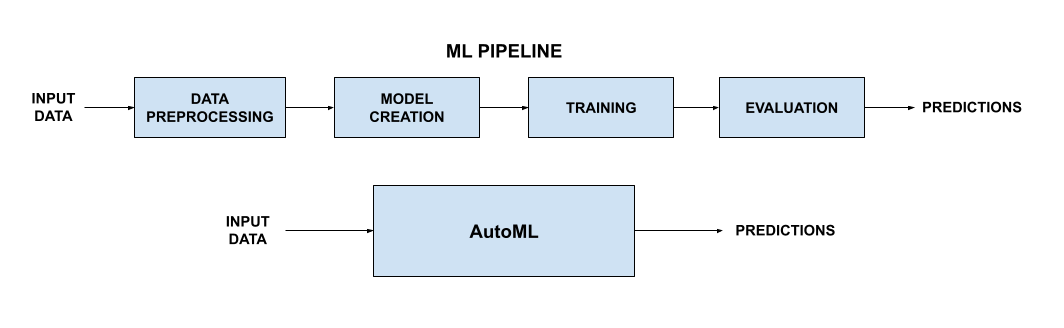}}
    \caption{AutoML Vs ML.}
    \label{fig:AutoML}
\end{figure}

\subsection{Tools and Platforms}
Every year more and more tools and platforms are emerging \cite{Gijsbers2019}. AutoML platforms are services, which are mainly accessible in the cloud. Therefore, for this task they are not preferred. Although when a cloud based MLOps platform selected, is possible  to have better compatibility. There are also libraries and API's written in python and c++,  which are much more preferable when an end-to-end cloud-based  MLOps platform has not been chosen. The ones stand out are Auto-Sklearn \cite{autosklearn}, Auto-Keras \cite{autokeras}, TPOT \cite{tpot}, Auto-Pytorch \cite{autopytorch}, BigML \cite{bigml}. The main platforms are Google Cloud AutoML \cite{googlecloud}, Akkio \cite{akkio}, H2O  \cite{h2o}, Microsoft Azure AutoML \cite{azure} and Amazon SageMaker Autopilot \cite{amazon}. The most important tools are listed in table IV.
 \begin{table}[h!]
  \begin{center}
    \label{tab:table4}   
    \begin{tabular}{c|c|c}
      \toprule 
      \textbf{Name} & \textbf{class} & \textbf{Status}\\
      \midrule 
      Auto-sklearn  & Tool & Open Source\\
      Auto-Keras & Tool & Open Source\\
      TPOT & Tool & Open Source\\
      Auto-Pytorch& Tool &Open Source \\
      BigML& Tool and Platform & commercial\\
      Google Cloud AutoML  & Platform & Open Source\\
      Akkio & Platform & Open Source\\
      H2O & Platform & Commercial\\
      Microsoft Azure AutoML &Platform  &commercial \\
      Amazon SageMaker Autopilot & Platform & commercial \\
      \bottomrule 
    \end{tabular}
    \caption{AutoML Tools And Platforms.}
  \end{center}
\end{table}
 
\subsection{Combining MLOps and AutoML}
It is obvious that the combination of the two techniques can be extremely effective \cite{Ruf2021}, but there are still some pros and cons. AutoML requires a vast computational power in order to perform. The development of technological means 
computational power but every year more power is getting closer and closer to overcome these kind of challenges, but still AutoML will always be more computational expensive compare to classic machine learning techniques, mostly because they perform the same tasks in  much more less time. Also, we are given much less flexibility. The AutoML tool works as a pipeline and so we have no control over the choices it will make. So AutoML does not qualify for very specialized tasks. On the other hand, with AutoML retraining is a much easier and straightforward task. As long as the new data are labeled or the models use unsupervised techniques, we only have to feed the new data to AutoML tool and deploy the new model. In conclusion, AutoML is a much more quicker and efficient process than the classic ML pipeline  \cite{Feurer2018}, which can be extremely beneficial in the achievement of efficient and high maturity level MLOps systems.

\section{MLOps Challenges}
In the past years, lots of research tends to focus on the maturity levels of MLOps and the transition to fully automated pipelines \cite{John2021}. Several challenges have been detected in this area and it is not always easy to overcome them \cite{Fursin2020}. A low maturity level system relies on the classical machine learning techniques and requires an extremely good connection between the individual working teams such as data scientists, ML engineers and frond end engineers. Lots of technical problems arise from this deviation and the lack of compatibility from one step to another. The first challenge lies in the creation of robust efficient pipelines with strong compatibility. Constant evolving is another critical point of a high maturity level of a MLOps platform, thus constant retraining shifts in the top of the current challenges.

\subsection{Efficient Pipelines}
A MLOps system includes various pipelines \cite{Zhou2020}. Commonly a \textit{data manipulation pipeline}, a \textit{model creation pipeline} and a \textit{deployment pipeline} are mandatory. Each of these pipelines must be compatible with the others, in a way that optimizes flow and minimizes errors. From this aspect it is critical to choose the right tools for the creation and connection of these pipelines. The shape of the targets determines the best combination of tools and techniques, whereat  you do not have an ideal combination for each problem, but the problem determines the combination to be chosen.  Also, it is always critical to use the same data preprocessing libraries in every pipeline. In this way, we will prevent the rise of multiple compatibility errors.

\subsection{Re-Training}
After monitoring and tracking your model performance, the next step is retraining your machine learning model \cite{Schelter2018}. The objective is to ensure that the quality of your model in production is up to date. However, even if the pipelines are perfect, there are many problems that complicate or even make retraining impossible. From our point of view, the most important of them is new data manipulation.
\subsubsection{New Data Manipulation}
When a model is deployed in production, we use new, raw data to make the predictions and use them to extract the final results. However, when we are using supervised learning, we do not have at our disposal the corresponding labels. So it is impossible to measure the accuracy and constantly evaluate the model. It is possible to perceive the robustness of the model only by evaluating the final results, which isn't always an option. Even if we manage to evaluate the model and find low metrics at new data, the same problem arises again. In order to retrain (fine tune) the model, the labels are prerequisites. Manually labeling the new data is a solution but slows down the process and fails at constant retraining tasks. An approach is using the trained model to label the new data or use unsupervised learning instead of supervised learning but also relies on the type of the problem and the targets of the task. Finally, there are types of data where there is no need for labeling. The most common area that uses this kind of data is time series and forecasting.

\subsection{Monitoring}
In most papers and articles, monitoring is positioned as one of the most important functions in MLOps \cite{Banerjee2020}. This is because to understand the results helps understanding the lack of the entire system. The last section shows the importance of monitoring not only for the accuracy of the model, but for every aspect of the system.
\subsubsection{Data monitoring}
Monitoring the data can be extremely useful in many ways. Detection of outliers and drift is a way to prevent a failure of the model and help the right training. Constant monitoring of the shape of the data is always opposed to training data it is away.
There are lots of tools and techniques for data monitoring and choosing the right ones also depends on the target.
\subsubsection{Model Monitoring}
Monitoring the accuracy of a model is a way to evaluate the performance in a bunch of data at a precise moment. For a high maturity level system, we need to monitor more aspects of our model and the whole system. In the previous years, lots of research \cite{Klaise2020}\cite{Tamburri2020} is focused on sustainability,  robustness \cite{Apostolidis2021}, fairness, and explainability \cite{Avramidis2022}. The reason is that we need to know more about the structure of the model, the performance, the reason why it works or it doesn't.

\section{Conclusion}
In conclusion, MLOps is the most efficient way to incorporate ML models in production. Every year more enterprises use these techniques and more research has been made in the area. But MLOps maybe has a different usage. In addition to the application of ML models in production, a fully mature MLOps system with continuous training can lead us to more efficient and realistic ML models. Further, choosing the right tools for each job is a constant challenge. Although there are many papers and articles for the different tools it is not easy to follow the guidelines and incorporate them in the most efficient way. Sometimes we have to choose between flexibility and robustness with the respective pros and cons. Finally, monitoring is a stage that must be one of the main points of  interest. Monitoring the state of the whole system using sustainability, robustness, fairness, and explainability is from our point of view the key for mature, automated, robust and efficient MLOps systems. For this reason, it is essential to develop model and techniques which enables this kind of monitoring such as explainable machine learning models. AutoML is maybe the game changer in the maturity and efficiency chase. For this reason, a more comprehensive and practical survey for the usage of AutoML in MLOps is necessary.  

\section*{Acknowledgment}
This work was supported by the MPhil program “Advanced Technologies in Informatics and Computers”, hosted by the Department of Computer Science, International Hellenic University, Kavala, Greece.
\bibliographystyle{IEEEtran}
\bibliography{mybib}{}
\end{document}